\documentclass[conference]{IEEEtran}
\IEEEoverridecommandlockouts
\usepackage{cite}
\usepackage{amsmath,amssymb,amsfonts}
\usepackage{algorithmic}
\usepackage{graphicx}
\usepackage{textcomp}
\usepackage{xcolor}

\usepackage{easy-todo}
\usepackage[caption=false]{subfig}
\usepackage[export]{adjustbox}
\usepackage{hyperref}

\newcommand\Tstrut{\rule{0pt}{2.6ex}}

\def\BibTeX{{\rm B\kern-.05em{\sc i\kern-.025em b}\kern-.08em
    T\kern-.1667em\lower.7ex\hbox{E}\kern-.125emX}}

\begin{document}

\definecolor{mygreen}{rgb}{0.01, 0.75, 0.24}


\newcommand{\kernn}[1]{\kern#1pt}
\newcommand{\raisee}[2]{\raisebox{#1}{\ensuremath{#2}}}
\newcommand{\colorr}[2]{\textcolor{#1}{#2}}

\newcommand{\mtiny}[1]{\mbox{\tiny\ensuremath{#1}}}
\newcommand{\mfootnotesize}[1]{\mbox{\footnotesize\ensuremath{#1}}}
\newcommand{\mscriptsize}[1]{\mbox{\scriptsize\ensuremath{#1}}}
\newcommand{\mlarge}[1]{\mbox{\large\ensuremath{#1}}}
\newcommand{\mlargee}[1]{\mbox{\Large\ensuremath{#1}}}
\newcommand{\mlargeee}[1]{\mbox{\LARGE\ensuremath{#1}}}
\newcommand{\mlargeeee}[1]{\mbox{\huge\ensuremath{#1}}}
\newcommand{\keraise}[3]{\kernn{#2}\raisee{#3pt}{\mtiny{#1}}}
\newcommand{\axstnn}[7]{\begin{array}{#7}\keraise{#1}{#2}{#3}\\\keraise{#4}{#5}{#6}\end{array}}
\newcommand{\bxstnn}[4]{\begin{array}[]{c}\mlargee{#1}\\[#4pt]\keraise{#2}{#3}{0}\end{array}}
\newcommand{\xconv}[5]{\kernn{-4}\axstnn{#4}{0}{0}{#3}{0}{0}{r}\kernn{-12}\bxstnn{\bb{C}}{#5}{8}{-18}
\kernn{-8}\axstnn{#2}{-2.5}{0}{#1}{-2.5}{0}{l}}
\newcommand{\xatt}[0]{\kernn{-6}\bxstnn{\bb{A}}{}{8}{-18}}
\newcommand{\xdense}[5]{\kernn{-6}\axstnn{#4}{0}{0}{#3}{0}{0}{r}\kernn{-14}
\bxstnn{\bb{F}}{#5}{16}{-17}\kernn{-17}\axstnn{#2}{5}{1}{#1}{2}{-1}{l}}
\newcommand{\xpool}[5]{\kernn{-6}\axstnn{#4}{0}{0}{#3}{0}{0}{r}\kernn{-10}
\bxstnn{\bb{P}}{#5}{5}{-15}\kernn{-15}\axstnn{#2}{0}{0}{#1}{0}{0}{l}}
\newcommand{\xinp}[5]{\kernn{-4}\axstnn{#4}{0}{0}{#3}{0}{0}{r}\kernn{-15}
\bxstnn{\cl{I}}{#5}{12}{-17}\kernn{-14}\axstnn{#2}{5}{1}{#1}{2}{-1}{l}}
\newcommand{\xdrop}[5]{\kernn{-2}\axstnn{#4}{0}{0}{#3}{0}{0}{r}\kernn{-15}
\bxstnn{\bb{D}}{#5}{2}{-17}\kernn{-18}\axstnn{#2}{0}{0}{#1}{5}{-1}{l}}
\newcommand{\xmerge}[5]{\kernn{-2}\axstnn{#4}{0}{0}{#3}{0}{0}{r}\kernn{-15}
\bxstnn{\bb{M}}{#5}{1}{-17}\kernn{-9.5}\axstnn{#2}{0}{0}{#1}{0}{-2}{l}}
\newcommand{\xgeneral}[5]{\kernn{-4}\axstnn{#4}{0}{0}{#3}{0}{0}{r}\kernn{-11}
\bxstnn{\bb{Q}}{#5}{0}{-18}\kernn{-9}\axstnn{#2}{-2.5}{0}{#1}{-2.5}{0}{l}}
\newcommand{\xunit}[3]{\kernn{-4}
\overset{#1}{\underset{\raisee{-1.5pt}{\mtiny{#2}}}{\bxstnn{\bb{U}}{#3}{2}{-18}}}}
\newcommand{\xunitdef}[3]{\xunit{#1}{#2}{}
\kernn{-4}\raisee{-2pt}{\ensuremath\longleftarrow\,\boxed{{#3}}}}
\newcommand{\xassign}[2]{{#1}\longleftarrow{#2}}
\newcommand{\xaggreg}[1]{\left\langle{#1}\right\rangle}
\newcommand{\xx}[2]{(#1)(#2)}
\newcommand{\bb}[1]{\mathbb{#1}}
\newcommand{\cl}[1]{\mathcal{#1}}
\newcommand{\tp}[1]{{#1}^{\intercal}}
\newcommand{\tr}[1]{\text{trace}\left[#1\right]}
\newcommand{\inv}[1]{\in\bb{R}^{#1}}
\newcommand{\inm}[2]{\in\bb{R}^{#1\times#2}}
\newcommand{\invc}[1]{\in\bb{C}^{#1}}
\newcommand{\inmc}[2]{\in\bb{C}^{#1\times#2}}
\def\ds{\displaystyle}
\def\ass{\leftarrow}
\def\od#1#2{\nabla_{#2}#1}
\def\tod#1#2{\tp{\nabla}_{#2}{#1}}
\def\cl#1{\mathcal{#1}}


\title{MinkLoc++: Lidar and Monocular Image Fusion for Place Recognition\\
\thanks{The project was funded by POB Research Centre for Artificial Intelligence and Robotics of Warsaw University of Technology within the Excellence Initiative Program - Research University (ID-UB)}
}

\author{\IEEEauthorblockN{Jacek Komorowski$^1$, Monika Wysoczańska$^1$, Tomasz Trzcinski$^{1,2}$}
\IEEEauthorblockA{\textit{$^1$Institute of Computer Science, Warsaw University of Technology \, $^2$Tooploox}\\
\{jacek.komorowski,monika.wysoczanska.dokt,tomasz.trzcinski\}@pw.edu.pl}
}

\maketitle

\begin{abstract}
We introduce a discriminative multimodal descriptor based on a pair of sensor readings: a point cloud from a LiDAR and an image from an RGB camera.
Our descriptor, named MinkLoc++, can be used for place recognition, re-localization and loop closure purposes in robotics or autonomous vehicles applications.
We use late fusion approach, where each modality is processed separately and fused in the final part of the processing pipeline.
The proposed method achieves state-of-the-art performance on standard place recognition benchmarks.
We also identify \emph{dominating modality} problem when training a multimodal descriptor. 
The problem manifests itself when the network focuses on a modality with a larger overfit to the training data. This drives the loss down during the training but leads to suboptimal performance on the evaluation set.
In this work we describe how to detect and mitigate such risk when using a deep metric learning approach to train a multimodal neural network.
Our code is publicly available on the project website.~\footnote{\url{https://github.com/jac99/MinkLocMultimodal}}
\end{abstract}


\section{Introduction}
\label{sec:intro}
Place recognition is an important task in computer vision with a broad range of practical applications, such as loop-closure in SLAM \cite{segmatch2017, zywanowski2020comparison} or autonomous driving \cite{s20102870, s18040939}. 
Early methods \cite{arandjelovic2016netvlad, doi:10.1177/0278364908090961,  6202705 , 7989362} mostly rely on images from RGB cameras and are affected by illumination changes and different weather conditions. 
To mitigate this problem and improve the robustness, autonomous vehicles are nowadays equipped with additional sensors, such as LiDAR.  LiDAR provides information about the geometry of the observed scene in the form of a sparse point cloud. It's more robust to adverse environmental conditions than RGB camera and can provide reliable data in limited visibility conditions (e.g. at night).
On the other hand, LiDARs cannot capture fine details of the observed scene, such as a texture of visible buildings or other objects, that might be helpful in localization task. 
The intuitive approach to improve place recognition performance is to combine these two modalities. Such multimodal approach is an interesting machine learning problem that has gained attention recently but remains underexplored~\cite{morency-baltrusaitis-2017-multimodal}.
The main research question is how to efficiently combine different modalities to fully exploit their complementary character.  

\begin{figure}
\centering
\includegraphics[width=1.0\linewidth,trim={0 7.8cm 12cm 0cm},clip]{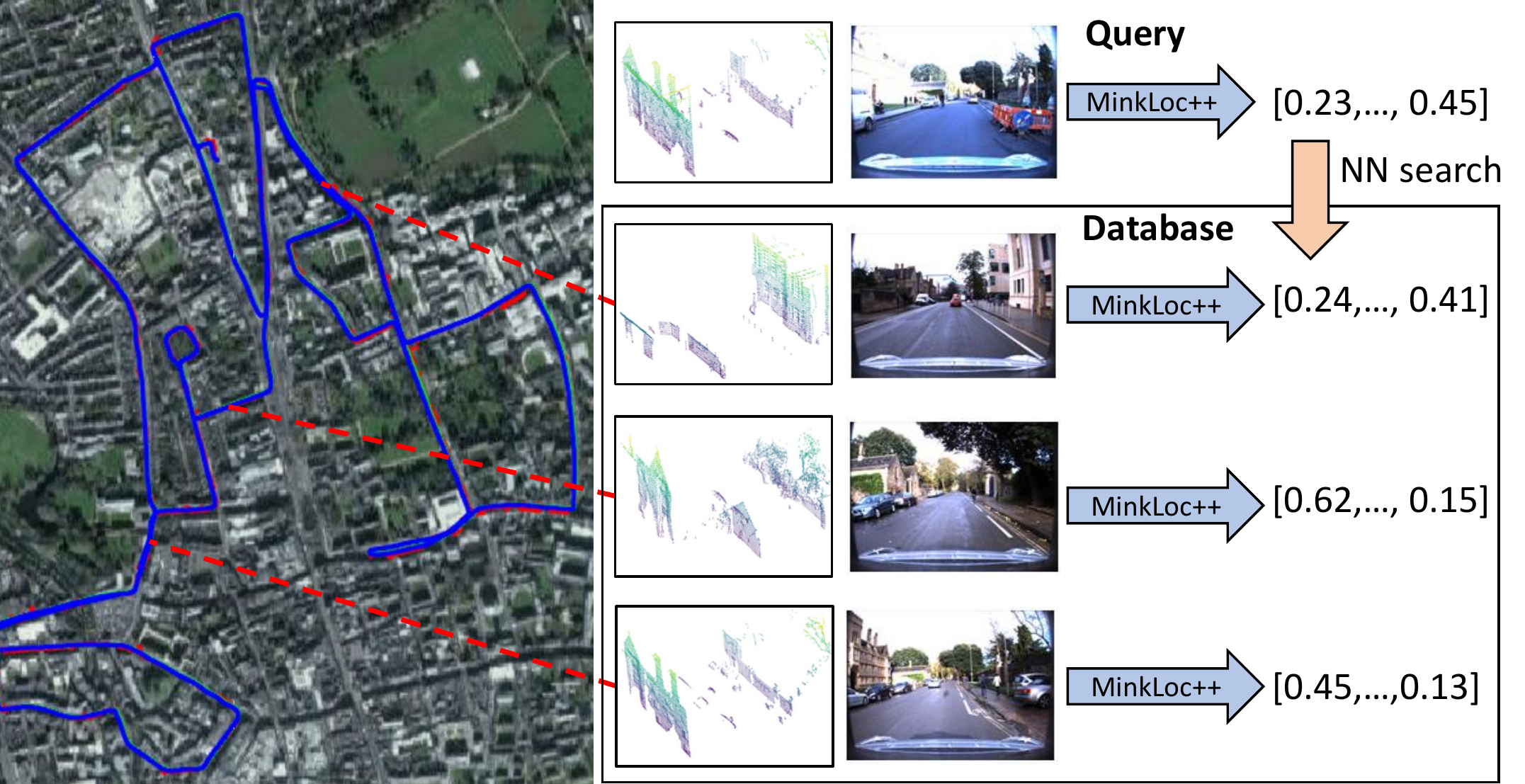}
\caption{Multimodal place recognition. MinkLoc++ computes a global descriptor from a pair of sensor readings: a 3D point  cloud from LiDAR and an image from a RGB camera.
Localization is performed by searching the database for a geo-tagged pair of readings with the closest descriptor.}
\label{fig:teaser}
\end{figure}

Place recognition is often formulated as an instance retrieval problem. Given a query instance, such as an RGB image, the aim is to find its closest match, with a known location, in a large-scale database. 
This idea is illustrated in Fig.~\ref{fig:teaser}.
In this work, we focus on learning a discriminative global descriptor for large-scale place recognition purposes using  multimodal data: an image from an RGB camera and a 3D point cloud from a LiDAR.
We use a \textit{late fusion} approach to obtain our multimodal descriptor, where two modalities are processed separately and merged in the final part of the processing pipeline. 
Our method yields the state-of-the-art results on the Oxford RobotCar \cite{RobotCarDatasetIJRR} and KITTI datasets \cite{Geiger2012CVPR}.

In our initial experiments we observed an interesting behaviour, when training the multimodal network using late fusion of two unimodal descriptors.
The network trained with two modalities performs worse compared to using only one modality.
Further investigation shows, that this counter-intuitive result is caused by different susceptibility to overfitting of two feature extraction subnetworks. 
The network learns to focus on the modality giving a smaller training error -- but because of much larger overfitting, producing poor results on the evaluation set. 
The same phenomenon is observed in the context of training multimodal classifiers in~\cite{wang2019makes}, where authors propose so-called Gradient-Blending approach based on overfitting-to-generalization ratio of each modality.
In this work we use deep metric learning approach with a triplet loss and propose a similar solution based on extending the loss function with additional terms based on individual modalities.

In summary the main contributions of our work are as follows.
\begin{itemize}
    \item We design a discriminative multimodal descriptor for place recognition purposes based on two modalities: a 3D point cloud from a LiDAR scan and an image from RGB camera. The proposed method advances state of the art.
    \item We propose an effective solution to the problem of \emph{dominating modality} which adversely affects the discriminability of a multimodal descriptor. The problem happens when one modality exhibits significantly higher overfitting to the training set than the other one.
\end{itemize}

The remainder of this paper is organized as follows: Sec.\ref{sec:related-work} gives a brief overview of the related work. Sec.\ref{sec:method} introduces our method for multimodal place recognition. Sec.\ref{sec:results} shows the experimental results and Sec.\ref{sec:conclusions} concludes our work. 

\section{Related work}
\label{sec:related-work}

\subsection{Image-based place recognition} 
The majority of image retrieval methods \cite{DBLP:journals/corr/abs-1906-07589,arandjelovic2016netvlad} builds upon local features aggregation techniques such as Fisher Vectors  \cite{sanchez:hal-00830491} or VLAD\cite{JDSP10}. Previously used handcrafted features are currently replaced with convolutional neural networks (CNNs), such as VGG \cite{7486599} or ResNet \cite{he2016deep}. 
In \cite{arandjelovic2016netvlad} authors introduce NetVLAD, an end-to-end learnable method explicitly targeting the large-scale place recognition task. Its main component is a generalized VLAD layer for local feature aggregation. It can be plugged into any CNN-based feature extractor and trained using backpropagation. Similarly, Generalized Mean Pooling (GeM) proposed in \cite{radenovic2018fine} is another aggregation method that can be appended a CNN backbone and trained end-to-end. GeM layer generalizes max and average pooling and proved successful in instance retrieval tasks. 

\subsection{Point cloud-based place recognition}
PointNetVLAD \cite{angelina2018pointnetvlad} is the first deep network to generate a global point cloud descriptor for large-scale place recognition. The method uses PointNet architecture \cite{qi2017pointnet} to extract local features, followed by NetVLAD aggregation layer.
However PointNet-based architecture is not well suited to capture local geometric structures and extract informative local features.
LPD-Net \cite{liu2019lpd} addresses this weakness by enhancing an input point cloud with handcrafted features and using graph neural networks to extract local contextual information.
EPC-Net \cite{hui2021efficient} improves upon LPD-Net by using proxy point convolutional neural network. 
SOE-Net \cite{xia2020soe} improves extraction of local contextual information by introducing the PointOE module, which captures local geometric structures from eight spatial orientations. 

All of the previously mentioned methods are based on unordered set of points representation. 
MinkLoc3D \cite{komorowski2020minkloc3d} uses an alternative representation and is based on sparse voxelized representation. It uses 3D convolutional architecture modelled upon a Feature Pyramid Network (FPN) \cite{DBLP:journals/corr/LinDGHHB16} design pattern to extract informative local features. Local features are aggregated using GeM~\cite{radenovic2018fine} layer into a discriminative global descriptor.
MinkLoc3D yields state-of-the-art results on standard benchmarks outperforming other other point cloud-based global descriptors.


\subsection{LiDAR and RGB camera fusion for multimodal place recognition}

Methods for multimodal place recognition can be categorized by a modality fusion strategy. 
The first group creates augmented images based on structural (3D) information, thus approaches modality fusion problem in the early stages of the processing pipeline. 
CORAL, a bi-modal descriptor presented in \cite{pan2020coral}, builds an elevation image from an input point cloud. 
Elevation image is enhanced with projected RGB image features extracted at multiple scales and processed using a deep neural network. Extracted local features are aggregated using NetVLAD layer.
In~\cite{zywanowski2020comparison} an input RGB image is concatenated with LiDAR intensity image and processed using a convolutional neural network.

The most common approach in multimodal place recognition is \textit{late fusion}.
In \cite{xie2020large} a point cloud descriptor is obtained using PointNetVLAD architecture, with NetVLAD aggregation layer altered to ignore non-informative 3D points.
Image-based descriptor is extracted using ResNet50~\cite{he2016deep} architecture.
Two unimodal descriptors are concatenated and processed by a fully-connected layer to produce the multimodal descriptor.
\cite{lu2020picnet} enhances \emph{late fusion} with a global channel attention layer (GCA). 
In \cite{9140362} image-based place recognition is augmented with structural cues constructed from RGB video stream using structure-from-motion. The reconstructed 3D point segments are discretized into a regular voxel grid and processed using 3D convolutional neural network. Features extracted from a 3D point cloud and an RGB image are merged to form the multimodal descriptor.

\section{Method description}
\label{sec:method}

This section presents our MinkLoc++ method to compute a discriminative global descriptor for place recognition purposes. The descriptor is based on two modalities: an image from an RGB camera and a point cloud from a LiDAR installed on a moving vehicle.
Place recognition is performed by searching the database of geo-tagged pairs of sensor readings (RGB image and 3D point cloud) for descriptors closest to the descriptor of the query pair. The idea is illustrated in Fig.~\ref{fig:teaser}.

\subsection{Network architecture}

\begin{figure*}
\centering
\includegraphics[scale=0.9,trim={0cm 0.2cm 0cm 0.2cm},clip]{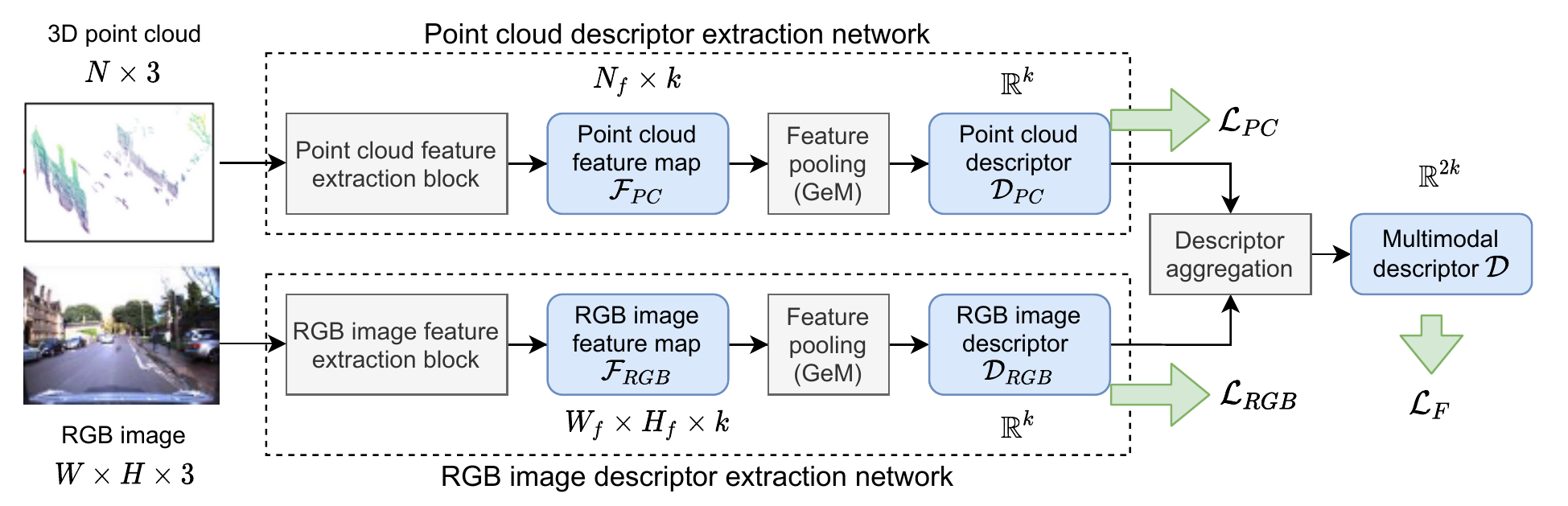}
\caption{High level MinkLoc++ architecture. Sensor readings (3D point cloud and RGB image) are processed by separate descriptor extraction networks, consisting of a local feature extraction block followed by a feature pooling layer.
Resultant point cloud descriptor $\mathcal{D}_{PC}$ and image descriptor $\mathcal{D}_{RGB}$ are aggregated to produce a fused multimodal descriptor $\mathcal{D}$. Green arrows show intermediary values used in the different terms of the loss function.
}
\label{fig:high_level}
\end{figure*}

We use \emph{late fusion} approach where modalities are processed separately and combined in the final part of the processing pipeline. The \emph{late fusion} approach is highly flexible and robust if one sensor fails. 

The high-level network architecture is shown in Fig.~\ref{fig:high_level}. 
The network consists of two branches: the upper branch computes a point cloud descriptor $\mathcal{D}_{PC} \in \mathbb{R}^k$ and the lower branch computes an RGB image descriptor $\mathcal{D}_{RGB} \in \mathbb{R}^k$. In our experiments we use $k=128$.

\begin{figure}
\centering
\includegraphics[scale=0.9,trim={0cm 0.2cm 0cm 0.1cm},clip]{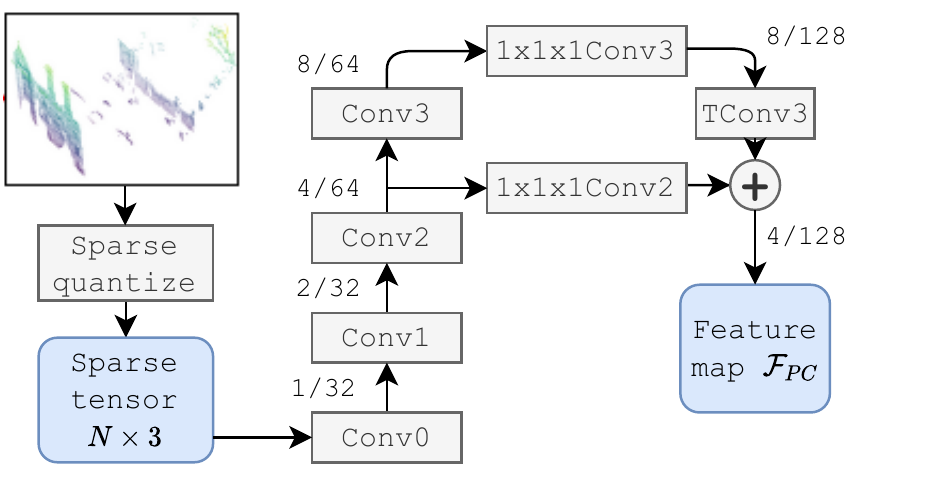}
\caption{Architecture of a \emph{point cloud feature extraction block}. 
The input point cloud is quantized into a sparse 3D tensor and processed by a 3D convolutional network with FPN~\cite{lin2017feature} architecture, producing a point cloud feature map $\mathcal{F}_{PC}$
Numbers in parentheses (e.g. 1/32) denote a stride and a number of channels of a feature map produced by each layer.
} 
\label{fig:cloud_fe}
\end{figure}

The upper branch uses an enhanced version of  MinkLoc3D~\cite{komorowski2020minkloc3d} architecture, that proved successful in point cloud-based place recognition task.
We improve the design by incorporating ECA~\cite{Wang_2020_CVPR} channel attention mechanism originally proposed for 2D convolutional networks.
The input point cloud is quantized into a sparse voxelized representation and processed by a  \emph{point cloud feature extraction block}, shown in Fig.~\ref{fig:cloud_fe}. 
It is a 3D convolutional network based on a Feature Pyramid Network~\cite{lin2017feature} design pattern. 
First, the input point cloud is processed in the bottom-up part direction producing 3D feature maps with decreasing spatial resolution and increasing receptive field. Each block in the bottom-up pass consists of a series of 3D convolutions with skip connections followed by  ECA~\cite{Wang_2020_CVPR} channel attention layer. See Fig.~\ref{tab:details_cloud_fe} for details.
Then, the transposed convolution upsamples the feature map generated by the last convolutional block. Upsampled feature map is concatenated with features from the corresponding layer of the bottom-up pass using a lateral connection. This design produces a feature map with relatively high spatial resolution and large receptive field. 
Our initial experiments proved, that pooling 3D features from feature map with higher spatial resolution is advantageous over using lower resolution maps computed by deeper networks.
The resultant sparse 3D feature map $\mathcal{F}_{PC}$ is pooled using a generalized-mean (GeM) pooling~\cite{radenovic2018fine} layer to produce a point cloud descriptor $\mathcal{D}_{PC} \in \mathbb{R}^k$. 

\begin{figure}
\begin{center}
\begin{tabular}{l@{\quad}l}
\hline
Block & Details
\\[2pt]
\hline
Conv0 & $\xconv{5_{k}1_{s}}{\ 32}{}{}{}$ 
\\
Conv1 &
$\xconv{2_{k}2_{s}}{\ 32}{}{}{}
\Big \langle \;
\xconv{3_{k}1_{s}}{\ 32}{}{}{} \;
\xconv{3_{k}1_{s}}{\ 32}{}{}{}
\xatt
\Big \rangle
$
\\
Conv2 & 
$\xconv{2_{k}2_{s}}{\ 64}{}{}{}
\Big \langle \;
\xconv{3_{k}1_{s}}{\ 64}{}{}{} \;
\xconv{3_{k}1_{s}}{\ 64}{}{}{}
\xatt
\Big \rangle
$
\\
Conv3 & 
$\xconv{2_{k}2_{s}}{\ 64}{}{}{}
\Big \langle \;
\xconv{3_{k}1_{s}}{\ 64}{}{}{} \;
\xconv{3_{k}1_{s}}{\ 64}{}{}{}
\xatt
\Big \rangle
$
\\
1x1x1Conv2, 3
&
$\xconv{1_{k}1_{s}}{\ 128}{}{}{}$
\\
1x1x1Conv3
&
$\xconv{1_{k}1_{s}}{\ 128}{}{}{}$
\\
TConv3 &  
$\xconv{2_{k}2_{s}}{\ 128}{t~}{}{}$
\\[5pt]
\hline
\end{tabular}
\end{center}

\caption{Layers in the \emph{point cloud feature extraction block}.
All convolutions in Conv$0\ldots3$ blocks are followed by batch norm and ReLU non-linearity.
$\mathbb{C}$ denotes a 3D convolution with a number of filters given as the top-right index, $t$ decorator indicates a transposed convolution, lower $k$ shows a filter size and lower $s$ a stride.
$\mathbb{A}$ is ECA~\cite{Wang_2020_CVPR} channel attention and 
$\langle \ldots \rangle$ enclosures a residual block with a skip connection.}
\label{tab:details_cloud_fe}
\end{figure}

The lower branch uses first four blocks of a pre-trained ResNet18~\cite{he2016deep} to extract a 2D feature map with 256 channels. The number of channels is reduced to $k$ using a convolutional layer with 1x1 filter, producing an RGB image feature map $\mathcal{F}_{RGB}$.
The  feature map is pooled using a GeM pooling to produce an RGB image descriptor $\mathcal{D}_{RGB} \in \mathbb{R}^k$.

\emph{Descriptor aggregation} block combines a point cloud descriptor $\mathcal{D}_{PC}$ and an image descriptor $\mathcal{D}_{RGB}$ to produce a final $2k$-dimensional multimodal descriptor. We use a simple concatenation along channels dimension to merge two descriptors. 
In ablation study we evaluate other options and discuss why this simplistic aggregation strategy gives better results than more sophisticated approaches.

\subsection{Network training}

To train the network to produce discriminative global descriptors we use a deep metric learning approach~\cite{lu2017deep} with a triplet margin loss~\cite{hermans2017defense}.
Each mini-batch element is a pair of readings: a 3D point cloud and a corresponding RGB image.
Mini-batch elements are arranged into triplets consisting of an anchor, a positive and a negative example. 
The positive example is a batch element similar to the anchor and a negative example is a batch element dissimilar to the anchor.
Two elements are similar if they show the same place, that is if they were acquired at locations at most 10 meters apart.
Elements are dissimilar, if they represent different locations, that is they were captured at locations more than 50 meters apart.
If a distance between locations is within 10-50 meters range, the elements are considered neither similar nor dissimilar. To increase the training efficiency, we use batch-hard negative mining~\cite{hermans2017defense} to build informative triplets.  

The loss function $\mathcal{L}$, defined by Eq.~\ref{eq:total_loss}, is a sum of three terms: the first is based on the fused multimodal descriptor $\mathcal{D}$, 
the second on the a point cloud descriptor $\mathcal{D}_{PC}$ 
and third on an RGB image descriptor $\mathcal{D}_{RGB}$.

\begin{equation}
\label{eq:total_loss}
\mathcal{L} = (1- \alpha - \beta) \mathcal{L}_F + \alpha \mathcal{L}_{PC} + \beta \mathcal{L}_{RGB},
\end{equation}
where $\alpha, \beta, \gamma$ are experimentally chosen weights and each component
$\mathcal{L}_F$, $\mathcal{L}_{PC}$, $\mathcal{L}_{RGB}$ is a triplet margin loss function~\cite{hermans2017defense} defined as:
\begin{equation}
\label{eq:loss}
\mathcal{L}_{\square}(a_i,p_i,n_i) = \max \left\{ d(a_i,p_i) - d(a_i,n_i) + m, 0 \right\} ,
\end{equation}
where 
\(d(x,y) = || x - y ||_2\) is an Euclidean distance between descriptors $x$ and $y$;  \(a_i, p_i, n_i\) are descriptors of an anchor, a positive and a negative element in $i$-th triplet and $m$ is a margin. 
For each batch, $\mathcal{L}_F$ is computed from triplets constructed from multimodal descriptors $\mathcal{D}$;  
$\mathcal{L}_{PC}$ from triplets constructed from point cloud descriptors $\mathcal{D}_{PC}$; 
and $\mathcal{L}_{RGB}$ from triplets constructed from RGB image descriptors $\mathcal{D}_{RGB}$.
Fig.~\ref{fig:high_level} shows the relation of each loss component to the processing pipeline.

The rationale for using multiple terms in the loss function is the \emph{dominating modality} problem discovered during our initial experiments.
We noticed that a multimodal descriptor, trained with a standard triplet loss, had an inferior performance on the evaluation set compared to the best unimodal descriptor.
This counter-intuitive result is caused by the fact that one modality (RGB image in our case) performs much better on the training set and much worse on the evaluation set than the other one (3D point cloud), due to much larger overfitting.
Thus, the network learns to focus on the modality with smaller training error, but greater evaluation error, largely ignoring the other one.
We mitigate this problem by extending the loss function with additional terms $\mathcal{L}_{PC}$ and $\mathcal{L}_{RGB}$ based on unimodal descriptors $\mathcal{D}_{PC}$ and $\mathcal{D}_{RGB}$ produced by two unimodal subnetworks (see Fig.~\ref{fig:high_level}).  This simple solution allows balancing the influence each modality and improves performance of the fused multimodal descriptor.

The same phenomenon was observed in the context of training multimodal classifiers in~\cite{wang2019makes}. 
It proposes Gradient-Blending, where weights of each loss term are based on overfitting-to-generalization ratio of each single modality subnetwork.
This approach cannot be directly applied when training a multimodal descriptor using a deep metric learning approach. When batch-hard negative mining is used, the loss function often fluctuates or stagnates for a number of epochs as harder and harder triplets are constructed. Instead, the number of active triplets (i.e. triplets with non-zero loss) constructed from each unimodal descriptor is much better indicator of overfitting. 
Hence, we find the dominating modality (RGB image) by examining the ratios of active triplets in training and validation set using each unimodal descriptor. Weights $\alpha$ and $\beta$ in Eq.~\ref{eq:total_loss} are chosen to maximize the performance on the validation set.
It should be noted, that it's sufficient to tune only one parameter in Eq.~\ref{eq:total_loss}, that is the weight for the non-dominating modality.
The weight for the dominating modality can be set to 0, as parameters of its descriptor extraction subnetwork will nevertheless be optimized during the learning due to their influence on the fused multimodal loss $\mathcal{L}_F$ term.
See ablation study section for evidence how multi-head loss improves discriminability of the multimodal descriptor.

The loss function is minimized using a stochastic gradient descent with Adam optimizer. 
Similar to~\cite{komorowski2020minkloc3d}, we use dynamic batch sizing.
The training starts with a small batch size (32 elements) and the batch size is gradually increased up to 128 elements as the training progress and the network learns more discriminative embeddings. 
The number of active triplets (i.e. triplets with non-zero loss) is monitored and when falls below the threshold, the batch size is increased.
This prevents the training process from collapse, which could happen in early epochs when mining hard triplets from a large batch, and allows using larger batch to mine more difficult triplets later on.

We use data augmentation to improve generalization and reduce the overfitting. 
For RGB images, we use photometric distortions, random crop and random erasing augmentation~ \cite{zhong2017random}.
For 3D points clouds, we use a random jitter and random points removal. We also adapted random erasing augmentation to remove 3D points within the randomly selected fronto-parallel cuboid.

\subsection{Implementation details}

In all experiments we quantize 3D point coordinates with 0.01 quantization step. As point coordinates in the Oxford RobotCar dataset are normalized to be within $\left[-1, 1\right]$ range, this gives up to 200 voxels in each spatial direction.
The initial learning rate for network parameters in the \emph{RGB image feature extraction block} is set to $10^{-4}$ and for all other parameters to $10^{-3}$.
In all experiments we train the network for 50 epochs, reducing the learning rate by 10 at the end of 30 epoch.
The dimensionality of point cloud and RGB image descriptors is set to $k=128$, and the multimodal descriptor has $2k=256$ dimensions.
To prevent embedding collapse in early epochs of training, we use a dynamic batch sizing strategy. The initial batch size is set to 8. When the number of active triplets falls below 70\% of the current batch size, the batch is increased by 40\% until the maximum size of 160 elements is reached.
To limit overfitting we use $L_2$ weight regularization with $\lambda=10^{-3}$ coefficient.
The coefficients of the loss terms in Eq.~\ref{eq:total_loss} are $\alpha=0.5, \beta=0$.

All experiments are performed on a server with a single nVidia RTX 2080Ti GPU, 12 core AMD Ryzen Threadripper 1920X processor and 64 GB of RAM. We use PyTorch 1.7~\cite{NEURIPS2019_9015} deep learning framework, MinkowskiEngine 0.4.3~\cite{choy20194d} auto-differentiation library for sparse tensors and Pytorch Metric Learning library 0.9.96~\cite{musgrave2020metric}.

\section{Experimental results}
\label{sec:results}

This section describes results of an experimental evaluation of our global descriptor and comparison with state of the art.

\subsection{Datasets and evaluation methodology}

Our method is trained and evaluated on a modified Oxford RobotCar dataset~\cite{RobotCarDatasetIJRR}.
The dataset is created using a suite sensors (RGB cameras, LiDAR sensors, GPS/INS) mounted on the car travelling multiple times through the same route in the city of Oxford at different times of day and year.
We build our dataset by enhancing a dataset of 3D point clouds introduced in~\cite{angelina2018pointnetvlad}, constructed from RobotCar LiDAR scans.
Point clouds are build from consecutive 2D LiDAR scans during the 20 m. drive of a vehicle. The ground plane is removed and point clouds are downsampled to 4096 points.
For each point cloud we find in an original RobotCar dataset corresponding RGB images with the closest timestamps taken from the center camera. Each image is downsampled from 1280x960 to 320x200 resolution.
During training, to increase data diversity and limit overfitting, we randomly sample from 15 closest RGB images. In evaluation we use only one RGB image with the closest timestamp.
Exemplary dataset elements are shown in Fig.~\ref{fig:teaser}.
Elements are similar (positive examples) if they are at most 10m apart and dissimilar (negative examples) if they are at least 50m apart.
Elements between 10 and 50m apart are neither similar nor dissimilar.
The dataset is split into disjoint training (21.7k elements) and test (3k elements) areas based on UTM coordinates.

We follow the same evaluation protocol and use the same train/test split  (so called Baseline scenario), as introduced in~\cite{angelina2018pointnetvlad} and used in later works~\cite{liu2019lpd,komorowski2020minkloc3d}.
An element (point cloud with a corresponding RGB image) from a test dataset is taken as a query. Pairs of point clouds and RGB images from other traversals covering the same region but captured at different times form the database. 
\emph{Recall@N} measures the percentage of correctly localized queries using top-$N$ elements returned from the database.
Localization is correct, if at least one of top-$N$ retrieved database elements is within 25 meter distance threshold from the ground truth position of the query element.
In addition to Average Recall@1 (AR@1) we report Average Recall@1\% (AR@1\%), which is calculated taking into account top-$k$ matches, where $k$ is 1\% of the database size.

Additionally, we evaluate the generalization ability of our methods using KITTI Odometry dataset. The dataset consists of RGB camera images, as well as LIDAR scans acquired with Velodyne laser scanner. Therefore, the characteristics of input point clouds differs significantly from the scans in Oxford RobotCar dataset. We follow the same procedure as introduced in \cite{pan2020coral}. We take Sequence 00 which visits the same places repeatedly and construct the reference database using the data gathered during the first 170 seconds. The rest is used as localization queries. 

\subsection{Evaluation results}

We compare the performance of our multimodal descriptor with point cloud-only descriptors: PointNetVLAD~\cite{angelina2018pointnetvlad}, PCAN~\cite{zhang2019pcan}, LPD-Net~\cite{liu2019lpd}, EPC-Net~\cite{hui2021efficient}, SOE-Net~\cite{xia2020soe}, MinkLoc3D~\cite{komorowski2020minkloc3d}
and multimodal descriptors based on a 3D point cloud and an RGB image: 
CORAL~\cite{pan2020coral}, PIC-Net~\cite{lu2020picnet}.
Where the source code or pretrained model is available (PointNetVAD, PCAN, LPD-Net, MinkLoc3D) we run the evaluation by ourselves, otherwise (EPC-Net, SOE-Net, CORAL, PIC-Net) we show results reported by authors on the same dataset and with identical evaluation protocol.

Evaluation results on the modified Oxford RobotCar dataset (Baseline scenario) are shown in Tab.~\ref{jk:tab:results_baseline}.
Our MinkLoc++ model outperforms other multimodal methods. 
The improvement in AR@1\% seems moderate (99.1 our method versus 98.2 for PIC-Net) but this metric is close to 100\% and there's very little room for improvement.
We also compared MinkLoc++ trained using only 3D modality to a number of unimodal, point cloud-based descriptors. Our method outperforms all of them, including recently proposed EPC-Net and SOE-Net. 
It can be noticed, that overall improvement from using multimodal approach compared to 3D-based unimodal is limited. The best unimodal descriptor has AR@1 94.5 compared to 96.7 obtained by multimodal MinkLoc++.
This can be explained by two factors. First, the results are close to 100\% and even 2 p.p. improvement is worth noticing. 
Second, RobotCar dataset was acquired at varying environmental conditions, with a number of images taken in limited visibility conditions (at night, very sunny or rainy day). 
3D modality based on LiDAR scans is more reliable source of data than an RGB image.

\begin{table}[htbp]
\caption{Evaluation results of place recognition methods on Robot Car dataset. 
}
\begin{center}
\begin{tabular}{l@{\quad}c@{\quad}r@{\quad}r}
\hline
\Tstrut
& \begin{tabular}{@{}c@{}}Modality \end{tabular}
& \begin{tabular}{@{}c@{}}AR@1\% \end{tabular}
& \begin{tabular}{@{}c@{}}AR@1 \end{tabular}
\\
[2pt]
\hline
\Tstrut
PointNetVLAD~\cite{angelina2018pointnetvlad} & 3D & 80.3 & 63.3 \\
PCAN~\cite{zhang2019pcan} & 3D & 83.8 & 70.7 \\
DH3D-4096~\cite{du2020dh3d} & 3D & 84.3 & 73.3\\
LPD-Net~\cite{liu2019lpd} & 3D & 94.9 & 86.4 \\
EPC-Net~\cite{hui2021efficient} & 3D & 94.7 & 86.2 \\
SOE-Net~\cite{xia2020soe} & 3D & 96.4 & 89.3 \\
MinkLoc3D~\cite{komorowski2020minkloc3d} & 3D & 97.9 & 93.8 \\
\textbf{MinkLoc++} (our) & 3D & \textbf{98.2} & \textbf{93.9} \\
[2pt]
\hline
\Tstrut
CORAL~\cite{pan2020coral} & 3D + RGB & 96.1 & 88.9 \\
PIC-Net~\cite{lu2020picnet} & 3D + RGB & 98.2 & - \\
\textbf{MinkLoc++} (our) & 3D + RGB & \textbf{99.1} & \textbf{96.7} \\
[2pt]
\hline
\end{tabular}
\end{center}
\label{jk:tab:results_baseline}
\end{table}

\captionsetup[subfigure]{labelformat=parens}

\begin{table}[htbp]
\caption{Evaluation results of place recognition methods on KITTI Odometry dataset.
}
\begin{center}
\begin{tabular}{l@{\quad}c@{\quad}r@{\quad}r}
\hline
\Tstrut
& \begin{tabular}{@{}c@{}}Modality \end{tabular}
& \begin{tabular}{@{}c@{}}AR@1\% \end{tabular}
\\
[2pt]
\hline
\Tstrut
PointNetVLAD~\cite{angelina2018pointnetvlad} & 3D & 72.4  \\
LPD-Net~\cite{liu2019lpd} & 3D & 74.6  \\
MinkLoc++ (our) & 3D & 72.6\\
MinkLoc++ (our) & RGB & 76.2 \\
[2pt]
\hline
\Tstrut
CORAL~\cite{pan2020coral} & 3D + RGB & 76.4 \\
MinkLoc++ (our) & 3D + RGB & \textbf{82.1}  \\
[2pt]
\hline
\end{tabular}
\end{center}
\label{jk:tab:results_kitti}
\end{table}

\begin{figure*}[htbp]
\centering
\subfloat{%
\includegraphics[width=4.33cm,height=3.1cm,trim={1.5cm 1cm 0.5cm 2cm},clip,cfbox=black 0.5pt 0.5pt]{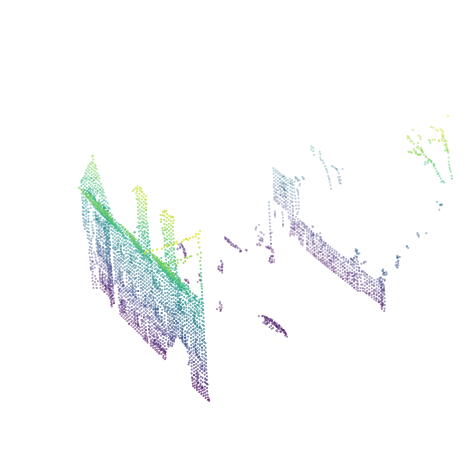}}
\hfill
\subfloat{%
\includegraphics[width=4.33cm,height=3.1cm,trim={1.5cm 1cm 0.5cm 2cm},clip,cfbox=red 0.5pt 0.5pt]{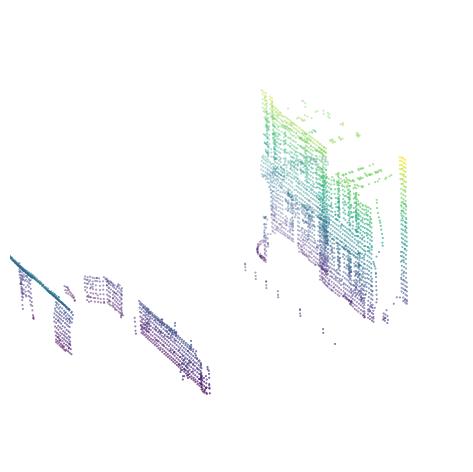}}
\hfill
\subfloat{%
\includegraphics[width=4.33cm,height=3.1cm,trim={1.5cm 1cm 0.5cm 2cm},clip,cfbox=red 0.5pt 0.5pt]{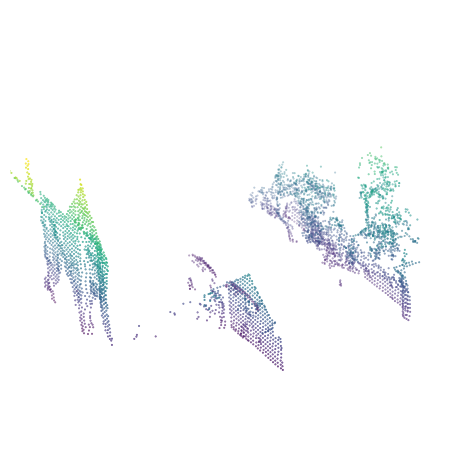}}
\hfill
\subfloat{%
\includegraphics[width=4.33cm,height=3.1cm,trim={1.5cm 1cm 0.5cm 2cm},clip,cfbox=mygreen 0.5pt 0.5pt]{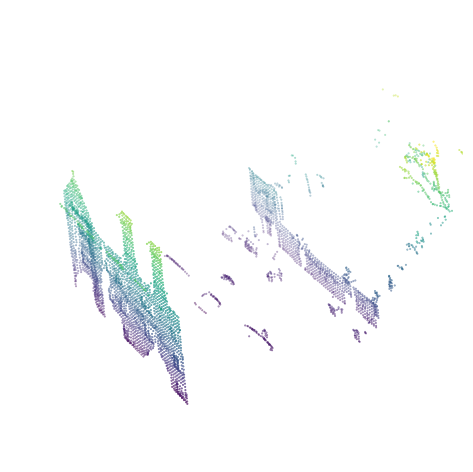}}
\\[-1ex]
\setcounter{subfigure}{0}
\subfloat[\label{11a}]{%
\includegraphics[width=4.33cm,height=3.1cm,trim={0cm 2cm 0cm 0cm},clip,cfbox=black 0.5pt 0.5pt]{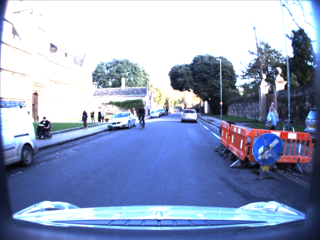}}
\hfill
\subfloat[\label{11b}]{%
\includegraphics[width=4.33cm,height=3.1cm,trim={0cm 2cm 0cm 0cm},clip,cfbox=red 0.5pt 0.5pt]{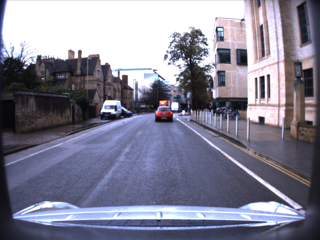}}
\hfill
\subfloat[\label{11c}]{%
\includegraphics[width=4.33cm,height=3.1cm,trim={0cm 2cm 0cm 0cm},clip,cfbox=red 0.5pt 0.5pt]{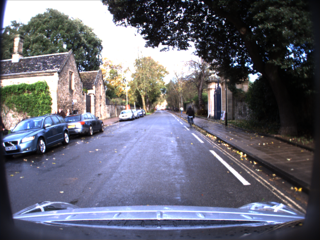}}
\hfill
\subfloat[\label{11d}]{%
 \includegraphics[width=4.33cm,height=3.1cm,trim={0cm 2cm 0cm 0cm},clip,cfbox=mygreen 0.5pt 0.5pt]{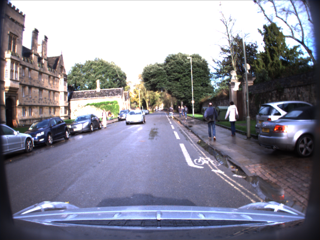}}
\caption{Example of successful retrieval using multimodal approach. 
(a) is the query element, (b) incorrect retrieval using 3D modality and (c) incorrect retrieval using RGB modality (d) correct retrieval using both modalities. 
}
\label{fig:multimodal_success}
\end{figure*}

\begin{figure}[htbp]
\centering
\setcounter{subfigure}{0}
\subfloat{\label{12a}%
\includegraphics[width=2.6cm,height=1.6cm,trim={11.5cm 1cm 0cm 2cm},clip,cfbox=black 0.5pt 0.5pt]{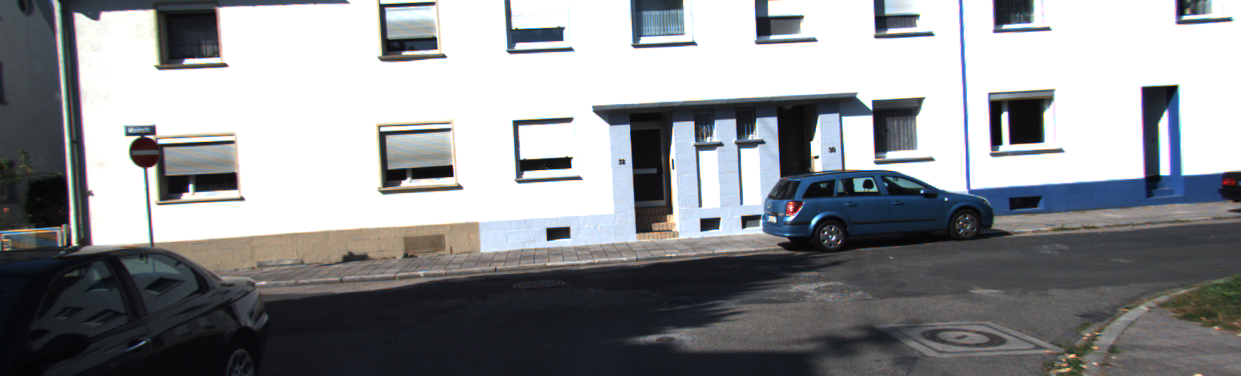}}
\hfill
\subfloat{%
\includegraphics[width=1.6cm,height=1.6cm,trim={1.5cm 0cm 0.5cm 0cm},clip,cfbox=black 0.5pt 0.5pt]{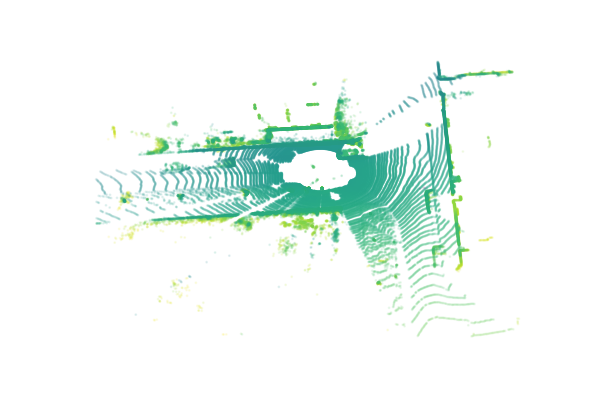}}
\hfill
\subfloat{\label{12b}%
\includegraphics[width=2.6cm,height=1.6cm,trim={0cm 1cm 11.5cm 2cm},clip,cfbox=mygreen 0.5pt 0.5pt]{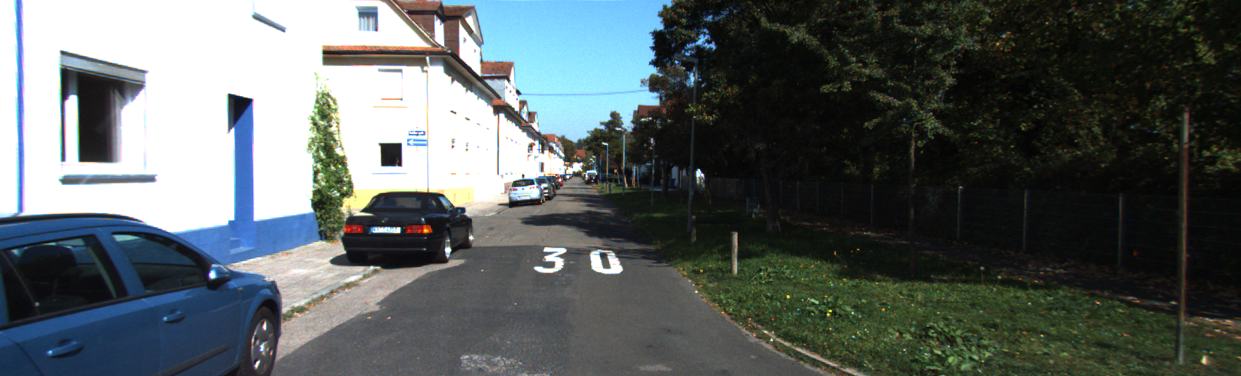}}
\hfill
\subfloat{%
\includegraphics[width=1.6cm,height=1.6cm,trim={1.5cm 0cm 0.5cm 0cm},clip,cfbox=mygreen 0.5pt 0.5pt]{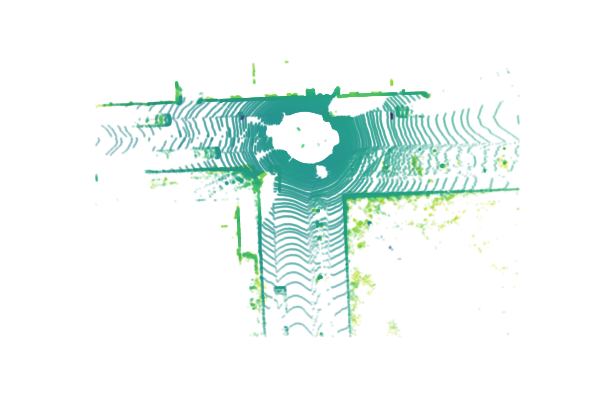}}
\\[-2ex]
\subfloat{\label{12c}%
\includegraphics[width=2.6cm,height=1.6cm,trim={10cm 1cm 1.5cm 2cm},clip,cfbox=red 0.5pt 0.5pt]{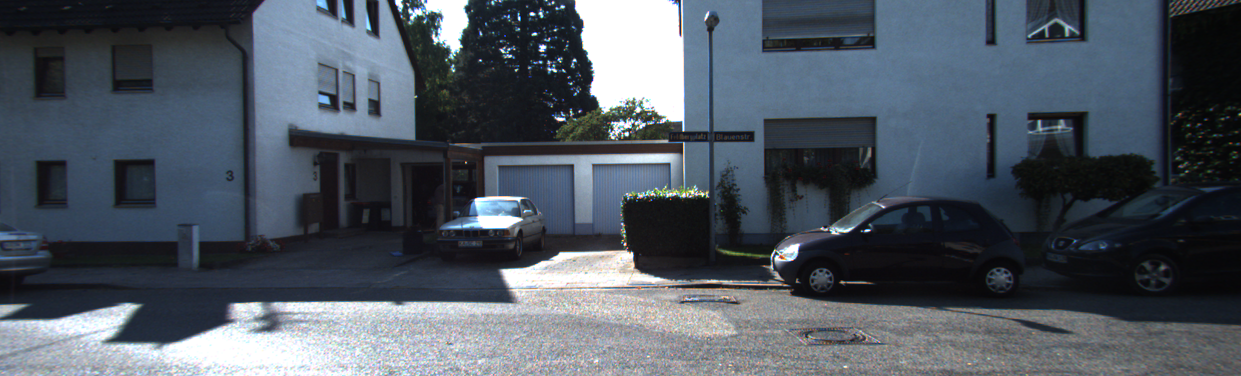}}
\hfill
\subfloat{%
\includegraphics[width=1.6cm,height=1.6cm,trim={1.5cm 1cm 0.5cm 0cm},clip,cfbox=red 0.5pt 0.5pt]{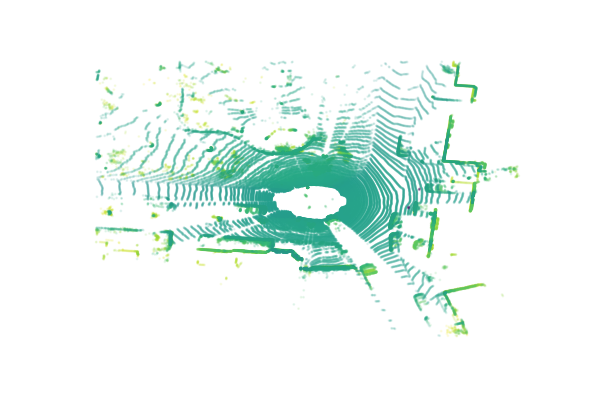}}
\hfill
\subfloat{\label{12d}%
\includegraphics[width=2.6cm,height=1.6cm,trim={10cm 1cm 1.5cm 2cm},clip,cfbox=red 0.5pt 0.5pt]{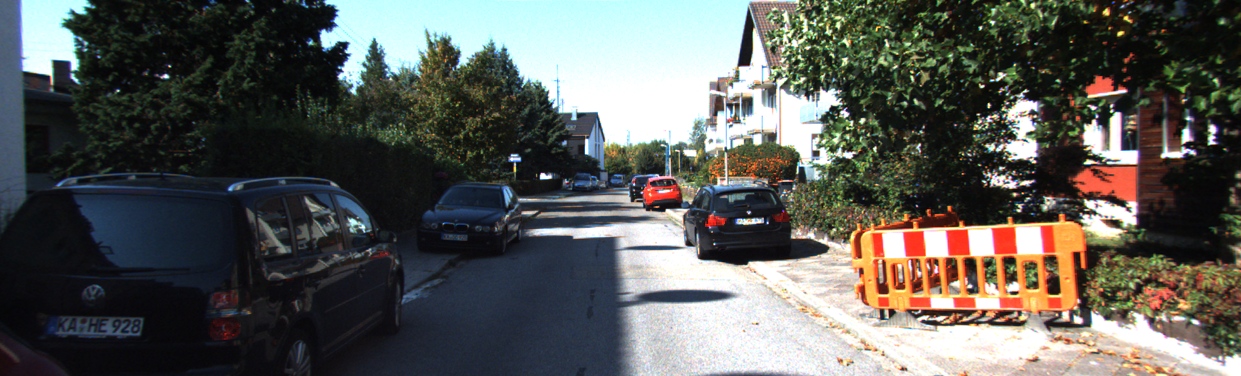}}
\hfill
\subfloat{%
\includegraphics[width=1.6cm,height=1.6cm,trim={1.5cm 1cm 0.5cm 0cm},clip,cfbox=red 0.5pt 0.5pt]{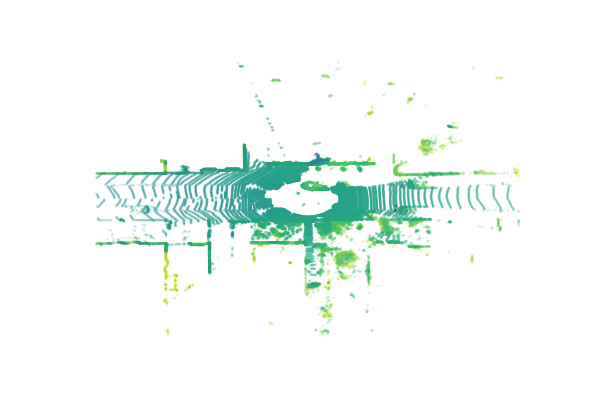}}
\caption{Example of successful retrieval using multimodal approach. 
The top left is the query example. The top right is a successful retrieval based on multimodal input. The bottom left is incorrect retrieval using RGB input, and the bottom right is the one based on 3D input.
}
\label{fig:multimodal_success_kitti}
\end{figure}

\begin{figure}[htbp]
\centering
\subfloat{%
\includegraphics[width=2.8cm,height=2.1cm,trim={1.5cm 1cm 0.5cm 2cm},clip,cfbox=black 0.5pt 0.5pt]{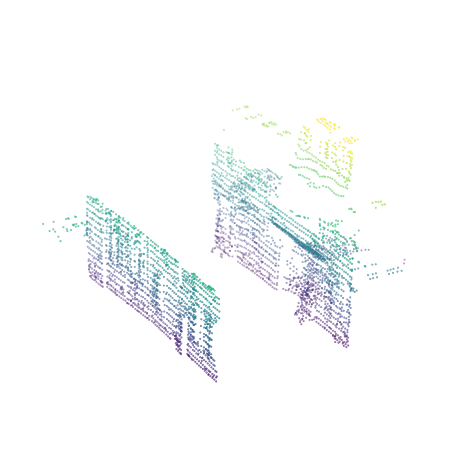}}
\hfill
\subfloat{%
\includegraphics[width=2.8cm,height=2.1cm,trim={1.5cm 1cm 0.5cm 2cm},clip,cfbox=red 0.5pt 0.5pt]{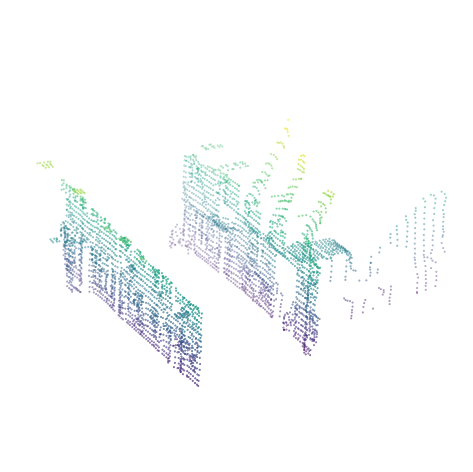}}
\hfill
\subfloat{%
\includegraphics[width=2.8cm,height=2.1cm,trim={1.5cm 1cm 0.5cm 2cm},clip,cfbox=mygreen 0.5pt 0.5pt]{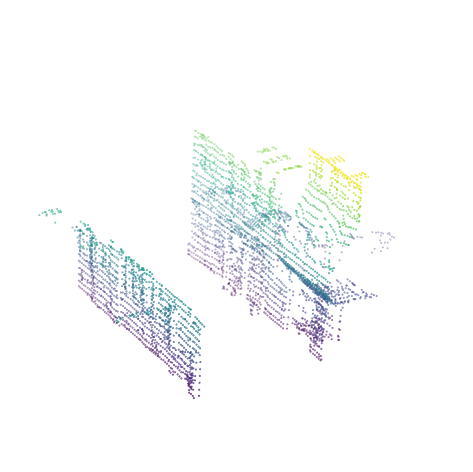}}
\\[-2ex]
\setcounter{subfigure}{0}
\subfloat[\label{13a}]{%
\includegraphics[width=2.8cm,height=2.1cm,trim={0cm 2cm 0cm 0cm},clip,cfbox=black 0.5pt 0.5pt]{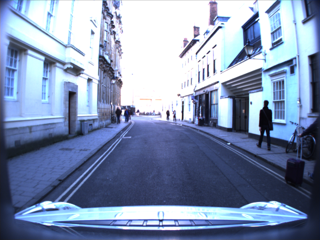}}
\hfill
\subfloat[\label{13b}]{%
\includegraphics[width=2.8cm,height=2.1cm, trim={0cm 2cm 0cm 0cm},clip,cfbox=red 0.5pt 0.5pt]{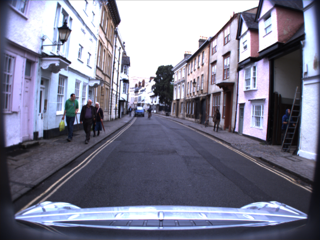}}
\hfill
\subfloat[\label{13c}]{%
 \includegraphics[width=2.8cm,height=2.1cm,trim={0cm 2cm 0cm 0cm},clip,cfbox=mygreen 0.5pt 0.5pt]{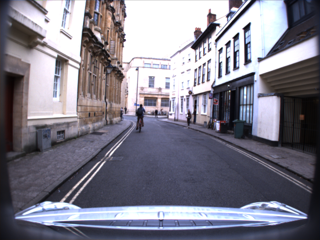}}
\caption{Example of unsuccessful retrieval using both modalities. (a) is the query element, (b) incorrect match to the query and (c) the closest true match.
Geometry of true and false matches is very similar and appearance of the query differs from the true match due to lighting conditions.}
\label{fig:failure_cases_multi}
\end{figure}

To verify robustness of our method to different environmental conditions,  we run additional evaluation on RobotCar Seasons~\cite{sattler2018benchmarking} benchmark. 
This is a challenging image-based dataset, prepared from Oxford RobotCar, with different splits corresponding to different atmospheric conditions and times of a day (e.g. snow, rain, dawn or night). 
We extend this dataset with 3D information to allow testing our multimodal descriptor.
For each image, we find LiDAR readings with corresponding timestamps in the in original RobotCar dataset~\cite{RobotCarDatasetIJRR} and use them to construct 3D point clouds. 
We compute descriptors for each element in the extended dataset and use them to retrieve elements from each dataset split.
We compare our method with two image-based and two point cloud-based global descriptors.
Tab.~\ref{jk:tab:results-seasons} reports the percentage of correctly localized queries for each dataset split.
The query is correctly localized if the pose of best matching element from the database is within 5 meter and 10$^{\circ}$ threshold from the ground truth pose.
In day conditions MinkLoc++ consistently outperforms all unimodal methods, except for rainy weather.
The rain adversely affects LiDAR performance causing worse results.
The improvement over image-based methods in good visibility conditions is only moderate and additional modality (3D) does not significantly improve the results.
In night conditions, point cloud-based methods and our multimodal method show their potential and overcome image-based methods by a large margin.

\begin{table*}[htbp]
\caption{Percentage of correctly localized queries on RobotCar Seasons benchmark.}
\begin{center}
\begin{tabular}{l@{\quad}|l@{\quad}|r@{\quad}r@{\quad}r@{\quad}r@{\quad}r@{\quad}r@{\quad}r@{\quad}|r@{\quad}r}
&  & \multicolumn{7}{c}{day conditions} & \multicolumn{2}{c}{night conditions} \\
& &  \begin{tabular}{@{}c@{}}dawn\end{tabular}
& \begin{tabular}{@{}c@{}}dusk\end{tabular}
& \begin{tabular}{@{}c@{}}overcast\\summer\end{tabular}
& \begin{tabular}{@{}c@{}}overcast\\winter\end{tabular}
& \begin{tabular}{@{}c@{}}rain\end{tabular}
& \begin{tabular}{@{}c@{}}snow\end{tabular}
& \begin{tabular}{@{}c@{}}sun\end{tabular}
& \begin{tabular}{@{}c@{}}night\end{tabular}
& \begin{tabular}{@{}c@{}}night-rain\end{tabular}
\\[2pt]
\hline
\Tstrut
DenseVLAD~\cite{torii201524} & RGB & 92.5 & 94.2 & 92.0 & 93.3 & \textbf{96.9} & 90.2 & 80.2 & 19.9 & 25.5 \\
NetVLAD~\cite{arandjelovic2016netvlad} & RGB & 82.6 & 92.9 & 95.2 & 92.6 & 96.0 & 91.8 & 86.7 & 15.5 & 16.4 \\
\hline
\Tstrut
LPD-Net~\cite{liu2019lpd}  & 3D & 79.7 & 79.9 & 79.7 & 73.8 & - & - & 82.3 & 77.3 & 32.8 \\
MinkLoc3D~\cite{komorowski2020minkloc3d} & 3D & 89.2 & 88.3 & 90.3 & 83.1 & 66.3 & 86.3 & 87.4 & \textbf{86.1} & 58.0 \\ 
[2pt]
\hline
\Tstrut
MinkLoc++ (our) & 3D+RGB & \textbf{93.6} & \textbf{94.4} & \textbf{96.3} &  \textbf{94.4} & 79.1 & \textbf{94.3} & \textbf{88.0} & 80.6 & \textbf{59.1} \\ 
[2pt]
\end{tabular}
\end{center}
\label{jk:tab:results-seasons}
\end{table*}

\subsection{Generalization evaluation}
To test the generalization ability of our method, we evaluate the model trained on RobotCar dataset on KITTI dataset. 
We compare our results with \cite{pan2020coral} by reproducing the evaluation procedure described in this paper.
Our multimodal MinkLoc++ descriptor outperforms other methods.
Our method can efficiently fuse two modalities to boost the performance. 
It must be noted, that point clouds in Kitti dataset have very different characteristic than point clouds in our training dataset.
The former are build from a single $360^{\circ}$ sweep of a 3D LiDAR, the latter are constructed by merging multiple scans from a 2D LiDAR during the 20m. drive of a vehicle. This explains worse performance of our method using solely 3D modality compared to the results on RobotCar dataset. 
Fig. ~\ref{fig:multimodal_success_kitti} shows an example of a successful retrieval case.


\subsection{Ablation study}
In this section, we investigate the effects of the design choices on our multimodal descriptor performance. In all experiments, unless otherwise noted, the network is trained using the same dataset and hyperparameters as described in previous sections.

First, we compare our multi-head loss function with a straightforward approach where the loss is based only on the fused multimodal descriptor.
Tab.~\ref{jk:tab:multi_head} shows the number of active triplets, at the end of the training, constructed from each unimodal descriptor and the performance of a fused multimodal descriptor, for different combination of weights in the loss function given by Eq.~\ref{eq:total_loss}. 
Setting  $\alpha$ and $\beta$ to 0 (first row), means that only one term $\mathcal{L}_{F}$, based on the fused multimodal descriptor, is used.
In this case, when using a training data there's much less ($-52$) active triplets for RGB image modality than for 3D modality.
On the contrary, on validation data, there's more ($+24$) active triplets for RGB modality than 3D. 
This suggests, that the network overfits much more to RGB modality, which is the \emph{dominating modality}.
Because of the larger overfit, during the training the network concentrates on RGB modality, largely ignoring the other one.
This leads to suboptimal performance of the fused multimodal descriptor on the evaluation set.
The solution is to increase the weight of the loss term corresponding to the non-dominating modality ($\mathcal{L}_{PC}$ in our case).
This balances both modalities during the training and increases performance of the fused multimodal descriptor.
Note, that the best performance of the multimodal descriptor is achieved when the difference between the number of active triplets for two unimodal descriptors is similar in both training and validation sets (rows 2 and 3 in Tab.~\ref{jk:tab:multi_head} with $\alpha=0.5$ and $\alpha=0.8$).
Using multi-head loss function allows increasing the performance from 94.4\% AR@1 (for $\alpha = \beta=0$) to 96.7\% (for $\alpha=0.5, \beta=0$).
In earlier experiments we tried limiting overfitting for RGB modality by regularization or more aggressive data augmentation. But this had only a limited impact and didn't allow solving the above mentioned problem. 

\begin{table}[htbp]
\caption{Impact of weights of each unimodal loss term in Eq.\ref{eq:total_loss} on the number of active triplets for each modality and the performance (AR\%1) of the multimodal descriptor.}
\begin{center}
\begin{tabular}{r@{\quad}r@{\quad}|r@{\quad}r@{\quad}c@{\quad}|r@{\quad}r@{\quad}c@{\quad}|r}
& & \multicolumn{3}{c}{Active triplets train}
& \multicolumn{3}{c}{Active triplets val}
&
\\
$\alpha$ & $\beta$ & \begin{tabular}{@{}c@{}} $\mathcal{D}_{PC}$ \end{tabular}
& \begin{tabular}{@{}c@{}} $\mathcal{D}_{RGB}$ \end{tabular}
&
& \begin{tabular}{@{}c@{}} $\mathcal{D}_{PC}$ \end{tabular}
& \begin{tabular}{@{}c@{}} $\mathcal{D}_{RGB}$ \end{tabular}
&
& \begin{tabular}{@{}c@{}}AR@1 \end{tabular}
\\ [2pt]
\hline
\rule{0pt}{1.2\normalbaselineskip}
 0 & 0 & 87 & 35 & \fbox{-52} & 95 & 119 & \fbox{+24} & 94.4 \\
 0.2 & 0 & 51 & 56 & +5 & 83 & 124 & +41 & 96.0 \\
 0.5 & 0 & 25 &  66 & +41 & 75 &  121 & +46 & \textbf{96.7} \\
 0.8 & 0 & 19 & 73 & +54 & 71 & 123 & +52 & 96.4 
\end{tabular}
\end{center}
\label{jk:tab:multi_head}
\end{table}

In Tab.~\ref{jk:tab:best_per_modality} we compare discriminativity of the multimodal descriptor with its unimodal counterparts.
To make a fair comparison we increase dimensionality of each unimodal descriptor to 256, the same size as the multimodal descriptor.
Using multimodal approach improves discriminability of the global descriptor (96.7 AR@1) compared to its unimodal counterparts (86.8 for RGB and 94.5 for 3D). 
As expected, an RGB-based descriptor has the worst performance. Oxford RobotCar is a challenging dataset and image acquisition conditions vary significantly between different traversals. The same place visited during a sunny day and at dusk or at night looks differently.
The improvement over 3D modality was more limited.

\begin{table}[htbp]
\caption{Impact of each modality on the discriminability of MinkLoc++ descriptor.}
\begin{center}
\begin{tabular}{l@{\quad}r@{\quad}r}
\hline
\Tstrut
\begin{tabular}{@{}c@{}}Modality\end{tabular}
& \begin{tabular}{@{}c@{}}AR@1\%\end{tabular}
& \begin{tabular}{@{}c@{}}AR@1\end{tabular}
\\
[2pt]
\hline
\Tstrut
RGB & 95.9  & 86.8 \\
3D point cloud &  98.3 &  94.5 \\
RGB + 3D & \textbf{99.1} & \textbf{96.7} \\
[2pt]
\hline
\end{tabular}
\end{center}
\label{jk:tab:best_per_modality}
\end{table}

Tab.~\ref{jk:tab:pooling_compare} shows the impact of a feature pooling method on the performance of each unimodal descriptor. 
The following methods are evaluated: maximum activations of convolutions (MAC)~\cite{2016particular},
sum-pooled convolutional features (SPoC)~\cite{7410507}, generalized-mean pooling (GeM)~\cite{radenovic2018fine}, NetVLAD~\cite{arandjelovic2016netvlad} and NetVLAD with Context Gating~\cite{miech17loupe} (NetVLAD-CG). 
For RGB modality, all examined methods, except for SPoC, performed similarly, with AR@1 varying between 86.1 and 86.9. 
For 3D modality, generalized-main pooling, with one trainable parameter, yields the best results (94.5 AR@1).
Hence, we choose GeM as a feature pooling method for both modalities.
It can be noted that more sophisticated feature aggregation methods (NetVLAD and NetVLAD with Context Gating) didn't improved the performance. 
This can be explained by the fact, that NetVLAD layer with a large number of trainable parameters (about 6 mln. in our configuration) increases overfitting which prevents an improvement on the evaluation set. 


\begin{table}[htbp]
\caption{Impact of a feature pooling method on a discriminability (AR@1) of unimodal descriptors.}
\begin{center}
\begin{tabular}{l@{\quad}r@{\quad}r@{\quad}r@{\quad}r@{\quad}r}
\hline
\Tstrut
& \begin{tabular}{@{}c@{}}SPoC\end{tabular}
& \begin{tabular}{@{}c@{}}MAC\end{tabular}
& \begin{tabular}{@{}c@{}}GeM\end{tabular}
& \begin{tabular}{@{}c@{}}NetVLAD\end{tabular}
& \begin{tabular}{@{}c@{}}NetVLAD-GC\end{tabular}
\\
[2pt]
\hline
\Tstrut
RGB & 76.9 & \textbf{86.9} & 86.8 & \textbf{86.9} & 86.1 \\
3D &  86.7 & 91.6 & \textbf{94.5} & 89.2 & 89.8 \\
[2pt]
\hline
\end{tabular}
\end{center}
\label{jk:tab:pooling_compare}
\end{table}

We also investigate different approaches for aggregation of two unimodal descriptors to form the multimodal descriptor.   
Two basic strategies are concatenation and summation of unimodal descriptors and the results are shown in the second column ('-') in Tab.~\ref{jk:tab:aggregation_compare}.
After this initial step, we optionally process the merged descriptor using a fully connected layer (column FC in the table) or two-layer MLP (column MLP).
Interestingly, simple concatenation or addition of unimodal descriptors gives the best results.
When concatenation is followed by a fully connected layer, AR@1 drops by 9 p.p. to 87.8.
This behavior can again be attributed to the \emph{dominating modality} problem. 
When a fully connected layer or MLP is used to process merged descriptors, the benefits or using multi-head loss function, with additional loss term for each modality, are lost.
Added layers learn to extract information from \emph{dominating modality} (RGB image in our case) from merged data and focus on it. 
Due to larger overfitting to the training set this leads to noticeably worse performance on the evaluation set.
It can be noted that adding 2-layer MLP drops AR@1 to 85.0 (86.61), which is comparable with the performance of the unimodal descriptor based on the dominating RGB modality (86.8 AR@1).

\begin{table}[htbp]
\caption{Impact of unimodal descriptors aggregation method on the discriminability (AR@1) of the multimodal descriptor. 
(-) only concatenate/add, FC = with a fully connected layer, MLP = with 2 layer MLP.}
\begin{center}
\begin{tabular}{l@{\quad}r@{\quad}r@{\quad}r@{\quad}r}
\hline
\Tstrut
& \begin{tabular}{@{}c@{}}-\end{tabular}
& \begin{tabular}{@{}c@{}}FC\end{tabular}
& \begin{tabular}{@{}c@{}}MLP\end{tabular}
\\
[2pt]
\hline
\Tstrut
Concatenation & \textbf{96.7} & 87.8 & 86.6 \\
Addition & \textbf{96.7} & 90.7 & 85.0 \\
[2pt]
\hline
\end{tabular}
\end{center}
\label{jk:tab:aggregation_compare}
\end{table}

\section{Conclusion}
\label{sec:conclusions}
In this paper we present MinkLoc++, a multimodal descriptor for place recognition purposes based on an input from a LiDAR scanner and RGB camera.
Experimental evaluation proves that the proposed method outperforms state of the art on standard benchmarks.
The performance of our method can be attributed to two factors. 
First, we use an efficient 3D convolutional network, based on sparse voxelized representation, for producing a discriminative point cloud descriptor. 
Enhancing MinkLoc3D~\cite{komorowski2020minkloc3d} architecture with a ECA channel attention~\cite{Wang_2020_CVPR} improves results on 3D modality.
Second, we show how to efficiently train multimodal descriptor avoiding \emph{dominating modality} problem.
Straightforward approach of computing the loss based only on the fused multimodal descriptor leads to suboptimal results.
This is caused by the fact, that one modality (RGB image in our case) performs better on the training set, and much worse on the validation set, due to larger overfitting. 
The network learns to focus on the modality with better training performance leading to worse results on the evaluation set.
We mitigate this problem by extending the loss function with additional terms based on unimodal descriptors computed in intermediary steps of the processing pipeline. 
This simple solution balances both modalities and improves performance of the fused descriptor.

\bibliographystyle{IEEEtran}
\bibliography{my-bib}

\end{document}